\PassOptionsToPackage{table}{xcolor}
\documentclass[10pt,letterpaper]{article}
\usepackage{alibaba-report}
\usepackage{arydshln,latexsym,wrapfig,float,mdframed}
\usepackage{graphicx}
\usepackage{chngcntr,ragged2e,makecell,pifont,colortbl,siunitx,xfp,xstring}

\newcommand{\ca}{\cellcolor{blue!45}}
\newcommand{\cb}{\cellcolor{blue!30}}
\newcommand{\cc}{\cellcolor{blue!15}}
\newcommand{\cd}{\cellcolor{blue!5}}
\definecolor{lightgray}{gray}{0.9}
\definecolor{best}{RGB}{255,235,190}
\definecolor{second}{RGB}{220,235,255}
\newcommand{\bestcell}[1]{\cellcolor{best}\textbf{#1}}
\newcommand{\secondcell}[1]{\cellcolor{second}\underline{#1}}
\definecolor{myGreen}{RGB}{0,150,0}
\definecolor{myRed}{RGB}{200,0,0}
\newcommand{\perf}[2]{%
    #1%
    \rlap{$
        \,_{\IfBeginWith{#2}{-}%
            {\color{myGreen}\text{\tiny{(#2)}}}%
            {\color{myRed}\text{\tiny{(#2)}}}%
        }
    $}%
}

\tcbset{paperbox/.style={
  colback=blue!5,
  colframe=blue!50!black,
  coltitle=white,
  colbacktitle=blue!60!black,
  boxrule=1.5pt,
  rounded corners,
  fonttitle=\bfseries,
  enhanced,
  attach boxed title to top left={yshift=-2mm,xshift=2mm},
  boxed title style={
    rounded corners,
    borderline west={0pt}{0pt}{white},
    borderline east={0pt}{0pt}{white},
    borderline north={0pt}{0pt}{white},
    borderline south={0pt}{0pt}{white},
  },
}}
\newtcolorbox{definitionbox}[1][]{paperbox,title=Definition,#1}
\definecolor{thm}{RGB}{69,53,193}
\newcounter{assump}[section]

\newtcolorbox{assumbox}[1][]{paperbox,breakable,
  before upper={\refstepcounter{assump}},#1}
\newtcolorbox{thmbox}[1][]{paperbox,breakable,
  colback=green!5,colframe=green!50!black,colbacktitle=green!60!black,#1}

\titleformat{\paragraph}[runin]{\normalsize\bfseries}{}{0pt}{}

\title{Why Steering Works:\\ Toward a Unified View of Language Model Parameter Dynamics}
\author{
Ziwen Xu\textsuperscript{1,2},
Chenyan Wu\textsuperscript{1},
Hengyu Sun\textsuperscript{1},
Haiwen Hong\textsuperscript{$\dagger$,2},
Mengru Wang\textsuperscript{1},\\
Yunzhi Yao\textsuperscript{1},
Longtao Huang\textsuperscript{2},
Hui Xue\textsuperscript{2},
Shumin Deng\textsuperscript{1},\\
Zhixuan Chu\textsuperscript{1},
Huajun Chen\textsuperscript{1},
Ningyu Zhang\textsuperscript{$\dagger$,1}\\[6pt]
{\reportaffiliationfont\mdseries
\textsuperscript{1}Zhejiang University,\quad
\textsuperscript{2}Alibaba Group}\\[4pt]
{\normalfont\textsuperscript{$\dagger$}Corresponding Author.}}
\newcommand{\shortpapertitle}{Why Steering Works: Toward a Unified View of Language Model Parameter Dynamics}
\newcommand{\codeurl}{https://github.com/zjunlp/EasyEdit/blob/main/examples/SPLIT.md}
\renewcommand{\paperlinks}{\par\vspace{3pt}\begingroup
  \sffamily\fontsize{7.5}{10}\selectfont
  \noindent\makebox[\linewidth][c]{\scalebox{.9}{\paperlinkentry{\codeurl}{Code}}}
  \par\endgroup}
\renewcommand{\authorcontributions}{}
\draftfalse
\AtBeginDocument{\hypersetup{pdfauthor={Ziwen Xu, Chenyan Wu, Hengyu Sun, Haiwen Hong, Mengru Wang, Yunzhi Yao, Longtao Huang, Hui Xue, Shumin Deng, Zhixuan Chu, Huajun Chen, Ningyu Zhang}}}

\begin{document}
\maketitle
\begin{abstract}
Methods for controlling large language models (LLMs), including local weight fine-tuning, LoRA-based adaptation, and activation-based interventions, are often studied in isolation, obscuring their connections and making comparison difficult. In this work, we present a unified view that frames these interventions as dynamic weight updates induced by a control signal, placing them within a single conceptual framework. Building on this view, we propose a unified preference-utility analysis. This analysis separates control effects into two components: preference, defined as the tendency toward a target concept, and utility, defined as coherent and task-valid generation. Both components are measured on a shared log-odds scale using polarity-paired contrastive examples. Across methods, we observe a consistent trade-off between preference and utility: stronger control increases preference while predictably reducing utility. We further explain this behavior through an activation manifold perspective, in which control shifts representations along target-concept directions to enhance preference, while utility declines primarily when interventions push representations off the model’s valid-generation manifold. Finally, we introduce a new steering approach \textbf{SPLIT} guided by this analysis that improves preference while better preserving utility.

\paperlinks
\end{abstract}
\section{Introduction}

\begin{wrapfigure}{r}{0.55\textwidth}
    \centering
    \vspace{-2em}
    \includegraphics[width=\linewidth]{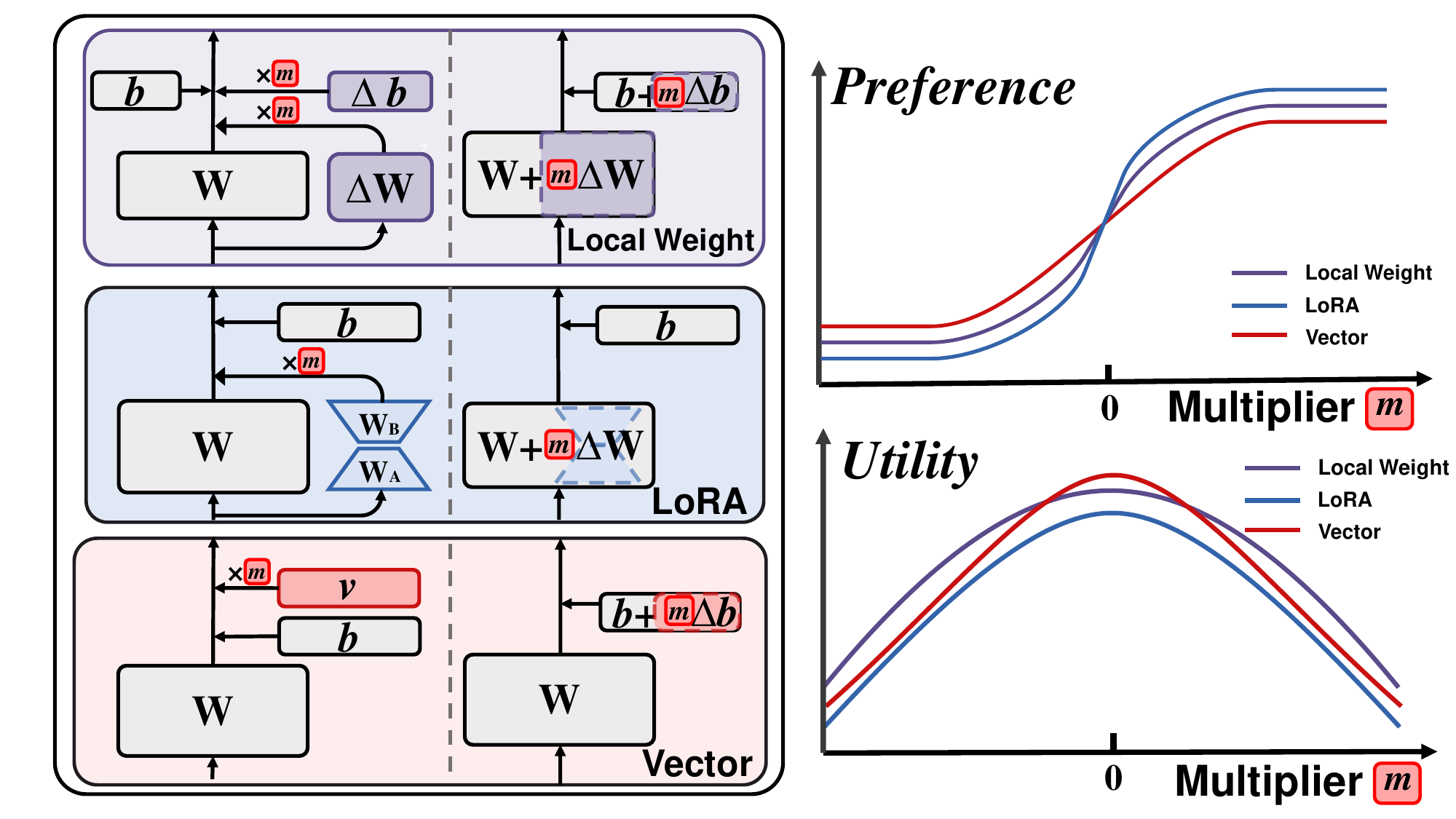}
    \caption{
        Diverse intervention methods operate on the model's linear layers and can be unified as \textbf{dynamic weight updates}.
        The right panel shows how utility and preference change with the intervention multiplier for each method (see Section~\ref{unified_view}).
    }
    \label{fig:steer}
    \vspace{-4em}
\end{wrapfigure}

Large Language Models (LLMs) have demonstrated remarkable capabilities and are increasingly deployed in real-world applications \cite{zhao2023survey}.
Growing demands for safety, controllability, and personalization make reliable control over model behavior a central challenge.
To address this, prior work has developed diverse paradigms for controlling LLMs, spanning training-time adaptations, such as local weight fine-tuning and parameter-efficient methods like LoRA \cite{hu2022lora,ding2023parameter,mao2025survey}, and inference-time interventions, including activation-level steering via hidden-state manipulation \cite{CAA,han2024word,DBLP:journals/corr/abs-2502-17601,DBLP:journals/corr/abs-2503-05613}.


Despite their empirical success, these approaches are often studied in isolation, under different assumptions, objectives, and evaluation protocols.
This fragmentation hinders rigorous comparison and obscures shared failure modes.
In this work, as shown in Figure \ref{fig:steer}, we mathematically observe that local weight fine-tuning, LoRA, and activation-level steering can all be formulated as instances of a common \emph{dynamic weight update} framework (Eq. \ref{eq:unified}).
Building on this perspective, we introduce a preference–utility analysis and show that, across methods instantiated within this framework, both preference and utility exhibit consistent, predictable patterns as control strength varies.

\begin{equation}
\mathbf{h}_{i+1} = (\mathbf{W} + m_1\,\Delta\mathbf{W})\,\mathbf{h}_i + (\mathbf{b} + m_2\,\Delta\mathbf{b}).
\label{eq:unified}
\end{equation}

Note that a particular challenge in controlled text generation is the \textbf{trade-off between enforcing the target concept and preserving task validity}: as control strength increases, the target attribute is amplified, but undesirable side effects—such as incoherence, instruction violations, or context drift—also become more frequent, reducing overall task effectiveness.
Moreover, because control quality is typically evaluated via realized outputs, degradation in task validity can confound assessments and obscure the intended concept signal.
Guided by this mechanistic understanding, we propose a training objective that explicitly optimizes preference while preserving utility, and experimentally demonstrate that it achieves superior performance.

Our contributions are as follows:
\begin{itemize}
    \item  \textbf{Unified View.} We propose a unified view of \emph{dynamic weight updates} that casts local weight fine-tuning, parameter-efficient fine-tuning (e.g., LoRA), and activation interventions (steering) into a common intervention form. 
    Building on this view, we introduce a unified preference--utility analysis and show that, across methods instantiated within the dynamic-update framework, both preference and utility exhibit consistent regularities as control strength varies.
    \item  \textbf{Preference–Utility Analysis.} We introduce an \emph{activation manifold} hypothesis and analyze preference and utility under this assumption, suggesting that preference is jointly determined by (i) the projection onto a target-preference direction and (ii) activation validity, which degrades as representations deviate from the manifold, while utility degradation is primarily driven by this off-manifold deviation and the resulting activation invalidation. 
    We further derive two quantitative relationships between the preference log-odds and $m$, and between the utility log-odds and $m$, and validate them with high-$R^2$ fits.
    \item  \textbf{New Steering Method.} Guided by this mechanism, we introduce \textbf{SPLIT}, a training objective  that explicitly optimizes preference while preserving utility, and demonstrate that it achieves better overall performance.
\end{itemize}

\section{Preliminary}
\label{sec:preliminary}
\subsection{Intermediate Representations in LLMs}
During the forward propagation of intermediate layers in LLMs, several key representations occur at specific points in the computation, such as FFN outputs, residual stream states, and linear projections within the attention mechanism (Query, Key, Value, and final output). These representations can be uniformly expressed as the output of an affine transformation:
\begin{equation}
\mathbf{h}_{i+1} = \mathbf{W} \mathbf{h}_i + \mathbf{b},
\label{eq:affine_form}
\end{equation}
where $\mathbf{h}_i$ and $\mathbf{h}_{i+1}$ denote the input and output representations of a linear layer, and $\mathbf{W}$, $\mathbf{b}$ are its weight matrix and bias vector, respectively.

For example, in an FFN block, the up-projection is
$\mathbf{h}_{\mathrm{mid}} = \mathbf{W}_{\mathrm{up}} \mathbf{h}_{\mathrm{in}} + \mathbf{b}_{\mathrm{up}}$,
followed by a non-linear activation, $\mathbf{h}_{\mathrm{mid,act}} = \sigma(\mathbf{h}_{\mathrm{mid}})$, and the down-projection is computed as
$\mathbf{h}_{\mathrm{out}} = \mathbf{W}_{\mathrm{down}} \, \mathbf{h}_{\mathrm{mid,act}} + \mathbf{b}_{\mathrm{down}}$.
Similarly, the $Q$, $K$, $V$, and output projections in the attention module follow the same affine form as in Eq.~\ref{eq:affine_form}.




\subsection{Parameter Update}
We consider two parameter adaptation methods for large language models: Low-Rank Adaptation (LoRA) and local weight fine-tuning.

\paragraph{LoRA} 
LoRA freezes the original weight matrix $\mathbf{W}$ and introduces a trainable low-rank update $\Delta\mathbf{W} = \mathbf{B}\mathbf{A}$, where $\mathbf{A} \in \mathbb{R}^{r \times k}$, $\mathbf{B} \in \mathbb{R}^{d \times r}$, and the rank $r \le \min(d, k)$. 
At inference, the adapted weights are given by $\mathbf{W} \leftarrow \mathbf{W} + \Delta\mathbf{W}$. 
In its canonical form, LoRA applies only to the weight matrix while keeping the bias term $\mathbf{b}$ fixed, although extensions exist that also adapt biases.

\paragraph{Local Weight Fine-tuning}
Local weight fine-tuning updates parameters within a restricted subset of the network, leaving all other parameters frozen. 
It can be applied to any layer or parameter type, with full-parameter training representing the special case where the subset covers the entire model. 
A generic update for the weight matrix $\mathbf{W}$ and bias vector $\mathbf{b}$ can be expressed as:
$(\mathbf{W}, \mathbf{b}) \leftarrow (\mathbf{W} + \Delta\mathbf{W},\; \mathbf{b} + \Delta\mathbf{b})$.
In our experiments, parameter updates are applied only to the \textbf{FFN down-projection} layer.

\subsection{Activation Steering}

\paragraph{Activation Steering} 
Activation steering modifies intermediate representations during inference by adding a steering vector to selected activations. 
Its mathematical form can be written as
\begin{equation}
\mathbf{h}_{i+1} = \mathbf{W} \mathbf{h}_i + \mathbf{b} + m \mathbf{v},
\end{equation}
where $\mathbf{v}$ is a predetermined direction and $m$ is a scalar coefficient controlling its magnitude. 
This approach builds on the \emph{linear representation hypothesis}~\citep{DBLP:conf/naacl/MikolovYZ13,DBLP:conf/emnlp/PenningtonSM14,DBLP:conf/blackboxnlp/NandaLW23,Tigges2023,DBLP:conf/icml/ParkCV24} that abstract concepts correspond approximately to linear subspaces of representation space.

The steering vector $\mathbf{v}$ can be equivalently expressed as a bias adjustment $\Delta\mathbf{b}$, yielding $\mathbf{b} \leftarrow \mathbf{b} + m \Delta\mathbf{b}$. 
This formulation highlights activation steering as a special case of dynamic parameter update, closely related to methods such as LoRA and local weight fine-tuning.

From a unified perspective, both parameter updates and activation steering operate by injecting a change vector $\Delta\mathbf{h}$ into intermediate representations during forward propagation, differing only in the mechanism by which $\Delta\mathbf{h}$ is generated.
More related works can be found in Appendix \ref{app:related}.

\section{Unified View of Dynamic Weights in Inference}
\label{unified_view}
We present a unified framework for dynamic interventions during inference, as illustrated in Figure~\ref{fig:steer}. 
Our unified view has three components: 
(i) a unified \emph{dynamic weights intervention} view that expresses local weight updates, LoRA, and activation steering as dynamic weight updates, 
(ii) a unified \emph{analysis} view based on preference/utility log-odds, and 
(iii) a unified \emph{dynamics observation} showing consistent preference--utility response patterns across intervention forms.

\begin{table*}[ht]
\centering
\begin{tabular}{lccc}
\toprule
\textbf{Form} & \textbf{Unified Affine Formula} & \textbf{Activation Impact} (\boldmath$\Delta\mathbf{h}$) & \textbf{Param. Size} \\
\midrule
Local Weight & $(\mathbf{W} + m\,\Delta\mathbf{W})\,\mathbf{h}_i + (\mathbf{b} + m\,\Delta\mathbf{b})$ & $m\,(\Delta\mathbf{W}\,\mathbf{h}_i + \Delta\mathbf{b})$ & $d_\text{in} \times d_\text{out} + d_\text{out}$ \\
LoRA & $(\mathbf{W} + m\,\mathbf{B}\mathbf{A})\,\mathbf{h}_i + \mathbf{b}$ & $m\,(\mathbf{B}\mathbf{A}\,\mathbf{h}_i)$ & $d_\text{in} \times r + r \times d_\text{out}$ \\
Steering Vector & $\mathbf{W}\,\mathbf{h}_i + (\mathbf{b} + m\,\Delta\mathbf{b})$ & $m\,\Delta\mathbf{b}$ & $d_\text{out}$ \\
\bottomrule
\end{tabular}
\caption{
All methods in our unified framework, expressed under the affine weight-update formulation and their corresponding activation changes $\Delta\mathbf{h}$. 
$d_\text{in}$ and $d_\text{out}$ denote the input and output dimensions of the layer; $r$ is the LoRA rank with $r \ll \min(d_\text{in}, d_\text{out})$.
}
\label{tab:unified_methods}
\end{table*}

\subsection{Unified Dynamic Weight Formulation}
\label{unified_formulation}
We propose a unified framework that encompasses both parameter update methods and activation steering methods, by viewing them as dynamic weight updates. 
Under this formulation, both can be expressed through a shared affine transformation view of intermediate representations; detailed derivations and formulations are provided in Section~\ref{sec:preliminary}.

Formally, the dynamic modification of the weight matrix $\mathbf{W}$ and bias vector $\mathbf{b}$ during inference is:
\begin{equation}
\mathbf{h}_{i+1} = (\mathbf{W} + m_1\,\Delta\mathbf{W})\,\mathbf{h}_i + (\mathbf{b} + m_2\,\Delta\mathbf{b}),
\label{eq:unified2}
\end{equation}
where $\Delta\mathbf{W}$ and $\Delta\mathbf{b}$ are update terms, and $m_1, m_2$ are scalar scaling coefficients controlling their magnitudes~\cite{DBLP:journals/corr/abs-2511-05408}. In other words, the original parameters are updated to $\mathbf{W}' = \mathbf{W} + m_1\,\Delta\mathbf{W}$ and $\mathbf{b}' = \mathbf{b} + m_2\,\Delta\mathbf{b}$ before computing the next-layer activation.

When a model weight is modified, the effect can be equivalently interpreted from the activation perspective, as a change to the activation at the corresponding position. 
In this view, diverse intervention methods are unified as adding a change term to the activation:
\begin{equation}
\Delta\mathbf{h} = m_1\,\Delta\mathbf{W}\,\mathbf{h}_i + m_2\,\Delta\mathbf{b}.
\end{equation}

Under this unified view, local weight fine-tuning, LoRA, and activation steering are all specific instances, differing only in which components are updated: 
local weight fine-tuning modifies both $\mathbf{W}$ and $\mathbf{b}$; 
LoRA modifies $\mathbf{W}$ via low-rank factors; 
activation steering modifies only $\mathbf{b}$. 
Table~\ref{tab:unified_methods} summarizes their affine forms, corresponding activation update, and parameter sizes. 

Notably, introducing explicit scaling coefficients extends traditional formulations and enables continuous control over perturbation strength, a capability that plays a central role in our subsequent analysis.

\subsection{Unified Analysis View: Preference and Utility Log-Odds}
\label{sec:logodds_formulation}

We analyze intervention effects along two complementary dimensions.
\textbf{Preference} denotes the model's internal inclination toward a target concept, independent of whether the model completion is well-formed.
For the prompt ``\emph{Write a short review for this restaurant}'', generating ``\emph{The food was excellent and the service was wonderful}'' indicates a positive preference, while ``\emph{The food was terrible and the service was disappointing}'' indicates a negative preference.
\textbf{Utility} denotes the model's task competence that is independent of the target concept.
It captures whether the model can produce a task-valid completion that is coherent, relevant to the prompt, and consistent with the requested format.
For the same prompt, utility is high when the output is a readable restaurant review, regardless of polarity.
Utility is low when the output is incoherent such as ``\emph{food food wonderful ??? service 19\% \#\#}'', off-topic such as ``\emph{Here is a Python script to scrape restaurant data...}'', or instruction-violating even if polarity-bearing words appear.

In controlled generation, performance is typically evaluated from the realized outputs.
When preference is increased at the expense of utility, completions often become incoherent or instruction-violating, reducing usability and obscuring the intended concept signal under output-based evaluation.
Therefore, effective model control should shift preference while preserving utility.

\paragraph{Notation.}
Given a query $q$, we construct a polarity pair of completions: a concept-positive answer $A_p$ and a concept-negative answer $A_n$.
We denote their conditional probabilities as $P(A_p\mid q)$ and $P(A_n\mid q)$, and define the corresponding cross-entropy losses as $\mathcal{L}_p \triangleq -\log P(A_p\!\mid q)$ and $\mathcal{L}_n \triangleq -\log P(A_n\!\mid q)$.
We further introduce latent preference probabilities $P(p_p\mid q)$ and $P(p_n\mid q)$, as well as a polarity-invariant task-success probability $P(u\mid q)$.

\paragraph{Preference--Utility Factorization.}
Following prior work that assumes concept directions are mutually orthogonal~\citep{belief-dynamics}, we
likewise treat concept preference as independent from task utility for a given query
$q$. Concretely, for a polarity pair $(A_p,A_n)$, we decompose
\begin{align}
P(A_p\mid q) &= P(u\mid q)\,P(p_p\mid q), \nonumber\\
P(A_n\mid q) &= P(u\mid q)\,P(p_n\mid q),
\end{align}
where $P(u\mid q)$ is shared across the pair and $P(p_p\mid q)+P(p_n\mid q)=1$.

\paragraph{Preference Log-odds.}
The shared utility cancels in the likelihood ratio, yielding
\begin{align}
\mathrm{PrefOdds}(q)
&\triangleq
\log\frac{P(p_p\mid q)}{P(p_n\mid q)}
=
\mathcal{L}_n-\mathcal{L}_p.
\label{eq:pref-odds_main}
\end{align}

\paragraph{Utility Log-odds.}
The total probability mass assigned to the matched pair recovers utility,
$P(u\mid q)=P(A_p\mid q)+P(A_n\mid q)$; substituting $P(A\mid q)=e^{-\mathcal{L}}$ gives
\begin{align}
\mathrm{UtilOdds}(q)
&\triangleq
\log\frac{P(u\mid q)}{1-P(u\mid q)}
=
\log\frac{e^{-\mathcal{L}_p}+e^{-\mathcal{L}_n}}
{1-e^{-\mathcal{L}_p}-e^{-\mathcal{L}_n}}.
\label{eq:util-odds_main}
\end{align}

We use $\mathrm{PrefOdds}$ and $\mathrm{UtilOdds}$ throughout to track how interventions shift concept preference versus task utility on a common additive scale, with additional derivations in Appendix~\ref{app:logodds}.


\subsection{Unified Dynamics Observation}
\label{sec:unified_observation}
\begin{figure*}[t]
    \centering
    \includegraphics[width=0.95\linewidth]{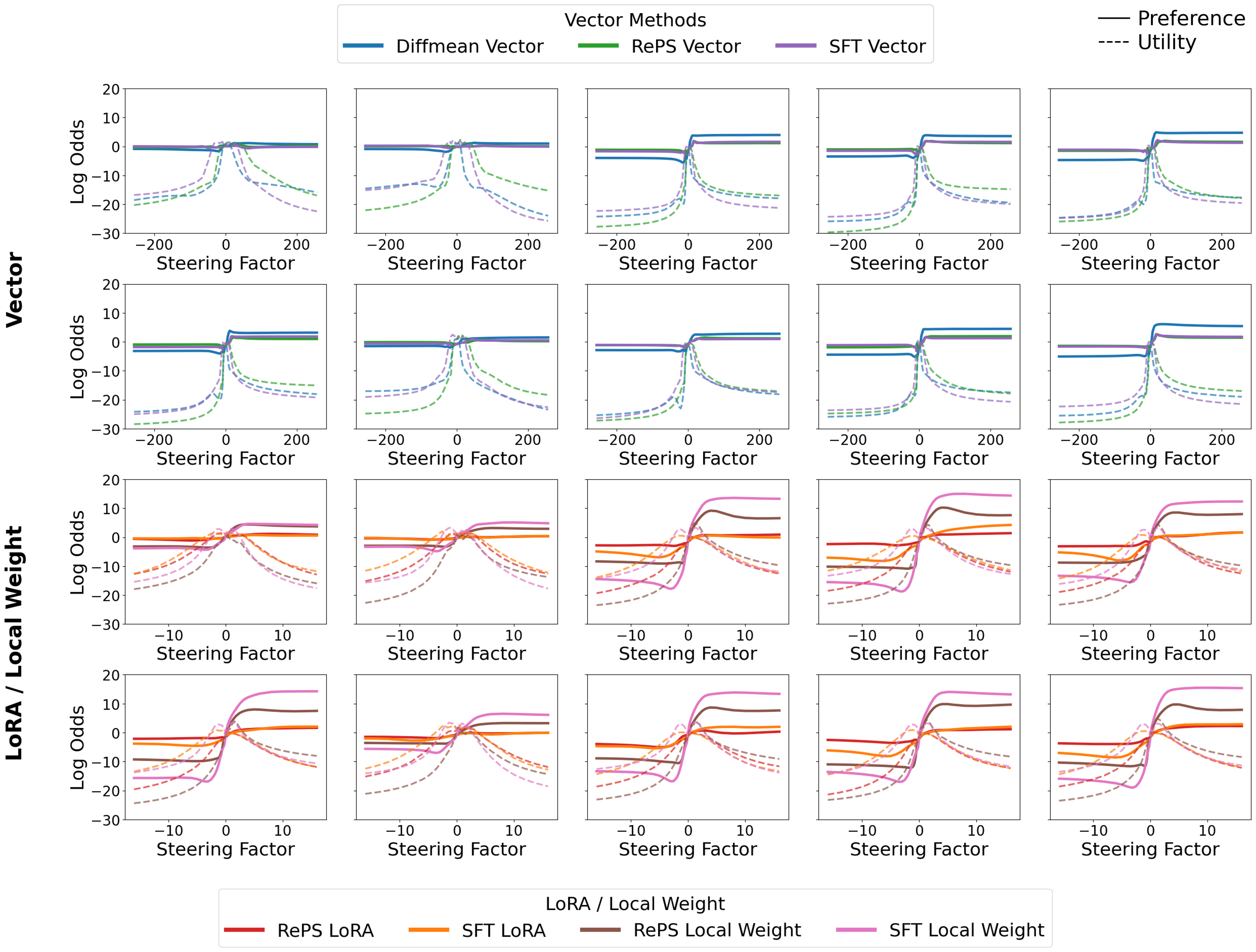}
    \caption{
        \textbf{Unified preference and utility dynamics under steering.}
        Solid lines represent preference log-odds, and dashed lines represent utility log-odds. 
        The top panel shows steering with vector-form parameter modifications, and the bottom panel shows parametric interventions including LoRA and local weight updates. 
        Results are shown for the \texttt{Gemma-2-9B-IT} model on the \emph{AxBench} dataset, evaluated over its top 10 concept subsets. 
        The horizontal axis corresponds to the steering factor.
    }
    \label{fig:unified-steering}
\end{figure*}

\paragraph{Experimental Setup.}
We evaluate dynamic interventions on two types of tasks: (i) a personality tendency classification task (\emph{Psychopathy}), and (ii) open-ended generation using \emph{PowerSeeking} and the top 10 concept subsets from \emph{AxBench}. 
We run experiments on \texttt{Gemma-2-9B-IT} at layer 20 and \texttt{Qwen-2.5-7B-Instruct} at layer 14, following~\citet{belief-dynamics}, and consider three intervention types: local weight, low-rank adaptation LoRA, and vector. 
We train each type using both the SFT objective and the RePS objective. 
Additionally, for vector, we include a train-free method called DiffMean~\cite{diffmean}. 
More details are provided in Appendix~\ref{experiment}.

\paragraph{Metrics.}
For each query $q$ with matched answers $(A_p,A_n)$, we compute preference and utility log-odds in Eqs.~\eqref{eq:pref-odds_main} and~\eqref{eq:util-odds_main}. 
These metrics track how preference and utility evolve as we vary the intervention scale $m$.

\paragraph{Unified Dynamics.} 
Experimental results show that, under the unified perspective framework, different intervention forms exhibit remarkably consistent dynamic patterns. 
As shown in Figure~\ref{fig:unified-steering}, localized weight updates, low-rank adaptation (LoRA), and vector-based interventions display highly similar overall curve shapes. 
Additional results are included in Appendix~\ref{experiment}.

For preference log-odds, all methods typically follow a three-stage pattern when plotted against the steering factor $m$: 
for small $|m|$, they enter a \emph{Linear Region}, where log-odds grows approximately linearly with $m$~\citep{belief-dynamics}; 
this is followed by a \emph{Transitional Region} with a noticeable change in trend, 
and finally a \emph{Convergence Region} where the curve flattens and stabilizes. 

Utility log-odds, in contrast, generally peak near $m \approx 0$, and remain near their maximum within this narrow range. 
As $|m|$ increases, utility gradually declines and eventually stabilizes.

These patterns reveal a unified steering response of preference and utility.

\section{Capability Dynamics: Mechanism Analysis and Optimization}
Motivated by the unified preference--utility dynamics observed across intervention forms (Figure~\ref{fig:unified-steering}), this section provides a mechanistic account and an empirical characterization. 
We take an activation-manifold perspective and introduce a simple validity-decay factor to capture the tendency for capability to degrade as steering pushes activations off the manifold, without committing to a specific underlying geometry. 
On this basis, we express preference as the combined effect of (i) steering-induced preference projection changes and (ii) validity decay, while utility is modeled as being dominated by the validity decay term. 
Finally, under this hypothesis we formalize how the steering factor $m$ shapes both preference and utility log-odds, and show via curve-fitting that the resulting forms match the observed log-odds--$m$ dynamics well across settings.

\begin{figure*}[t]
    \centering
    \includegraphics[width=0.9\linewidth]{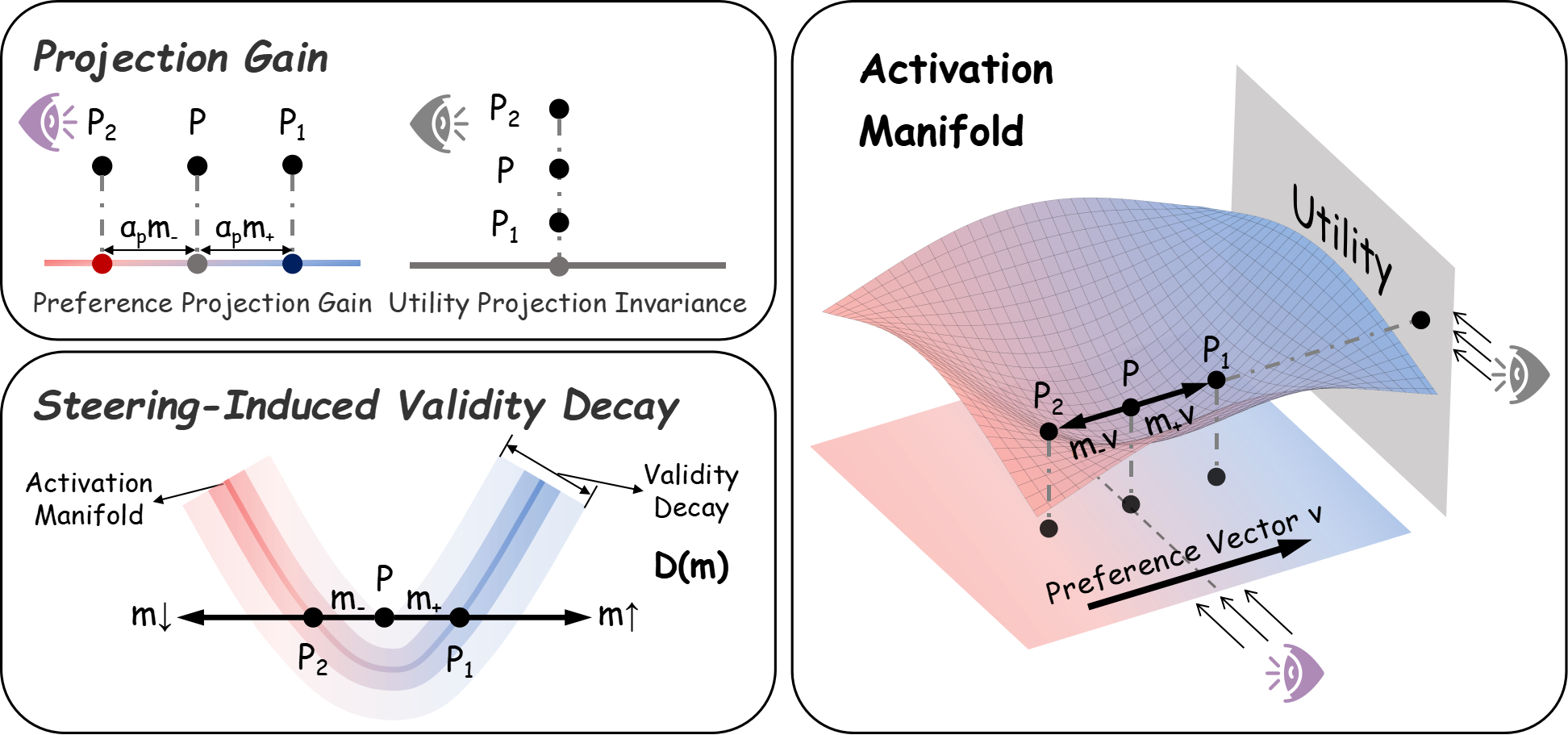}
    \caption{
        \textbf{Mechanism of projection gain and validity decay.}
        \textbf{Right:} An activation manifold view illustrating Assumption~\ref{ass:manifold}.
        An activation $P$ lies on or near the manifold. 
        Steering using preference vector $v$ with scaling factors $m_+$ and $m_-$ moves $P$ to $P_1$ and $P_2$, corresponding to intersections with the manifold.
        \textbf{Top-left:} Projection gain. 
        Projections onto the utility axis exhibit limited variation, whereas projections along the preference direction differ between $P_1$ and $P_2$, suggesting that steering primarily influences preference-related components.
        \textbf{Bottom-left:} Steering-induced validity decay. 
        As assumed in Assumption~\ref{ass:decay}, increasing steering factor increases off-manifold deviation, leading to a monotonic decrease in validity and degraded downstream decoding.
    }
    \label{fig:mechanism}
\end{figure*}

\subsection{Activation Manifold Hypothesis}
\label{sec:activation_manifold}
Prior work suggests that model activations often concentrate on low-dimensional, manifold-like sets in representation space~\citep{bricken2023monosemanticity,DBLP:conf/icml/WollschlagerEGC25}. 
Adopting this manifold perspective, we analyze additive steering as a translation of hidden states along an approximately fixed direction in activation space. 
Intuitively, small translations may adjust model behavior in a targeted way, whereas large translations may push representations away from the high-density region learned during training, increasing the risk of a representation--decoder mismatch and thus degrading general capability.

We formalize this view with two assumptions.
Assumption~\ref{ass:manifold} asserts that pre-training induces a “typical” region of activation space where representations concentrate for stably handled inputs. 
We next introduce a generic notion of \emph{representation validity}, which is high near $\mathcal{M}_l$ and decreases as hidden states move away from it.
This abstraction avoids committing to a specific geometry for $\mathcal{M}_l$ while retaining the key implication: sufficiently off-manifold activations are more likely to be decoded unreliably by the remaining network.

\begin{assumbox}
\paragraph{Assumption 4.1 (Training-Induced Activation Manifold).}
\label{ass:manifold}
Fix a layer $l$ with hidden dimension $d_l$. 
There exists a low-dimensional set (or its neighborhood) $\mathcal{M}_l \subset \mathbb{R}^{d_l}$ such that for inputs $x$ drawn from \emph{stably handled} inputs $\mathcal{X}_{\mathrm{stable}}$, the activation $h_l(x)$ lies on or near $\mathcal{M}_l$ with high probability:
\begin{equation}
\Pr_{x\sim \mathcal{X}_{\mathrm{stable}}}\!\left[d\!\left(h_l(x),\mathcal{M}_l\right)\le \epsilon\right]\ge 1-\delta,
\end{equation}
where $d(\cdot,\mathcal{M}_l)$ denotes distance to $\mathcal{M}_l$, $\epsilon>0$ is a radius, and $\delta\in(0,1)$ is a failure probability.
\end{assumbox}

\begin{assumbox}
\paragraph{Assumption 4.2 (Steering-Induced Validity Decay).}
Let $F_{l\rightarrow L}$ denote the remainder of the model from layer $l$ to the output logits. There exists a \emph{validity} function $V_l:\mathbb{R}^{d_l}\rightarrow[0,1]$ that is non-increasing in $d(h,\mathcal{M}_l)$, capturing how well $F_{l\rightarrow L}$ can stably decode an activation $h$.

For an additive steering intervention at layer $l$,
\begin{equation}
\tilde{h}_l(m)=h_l+m\,\Delta h,
\end{equation}
with direction $\Delta h$ and steering factor $m\in\mathbb{R}$, define the mean validity over stably handled inputs:
\begin{equation}
D(m)\;\triangleq\;\mathbb{E}_{x\sim \mathcal{X}_{\mathrm{stable}}}\!\left[V_l\!\left(\tilde{h}_l(m)\right)\right].
\end{equation}
We assume that $D(m)\in[0,1]$ decreases with $|m|$ (i.e., larger interventions induce larger off-manifold shifts on average), and that the resulting capability degradation is dominated by this validity decay.
\label{ass:decay}
\end{assumbox}

To connect Assumptions~\ref{ass:manifold}--\ref{ass:decay} to a concrete functional form, we view steering as moving an activation along a one-dimensional line in representation space, $\tilde h_l(m)=h_l+m\,\Delta h$. 
Under the manifold hypothesis, degradation is governed primarily by how far this line trajectory departs from the typical region near $\mathcal{M}_l$, so it is natural to model $D(m)$ as a smooth function of the (signed) distance along this line to the nearest ``on-manifold'' locations. 
In particular, as illustrated in Fig.~\ref{fig:mechanism}, the steered trajectory may intersect the manifold neighborhood at one or more values $\{m_i\}$ (e.g., one for $m>0$ and one for $m<0$). 
We therefore model validity as being highest near these intersection points and decaying as $|m-m_i|$ grows.

A convenient choice that is positive, smooth, and exhibits heavy-tailed distance-based decay is the rational quadratic (RQ) form, widely used in kernel methods and Gaussian processes to model multi-scale, polynomial-rate attenuation with distance~\citep{Rasmussen2004}. 
Prior work on controllability metrics has established that model steerability is often asymmetric~\cite{miehling-etal-2025-evaluating}, exhibiting varying degrees of responsiveness along different directions of the same dimension. Motivated by this, we employ a piecewise parameterized model to quantify degradation:
\begin{equation}
D(m)=\begin{cases} 
\left(1 + \frac{(m - m_{+})^2}{L_+}\right)^{-p_+} & \text{if } m \ge 0 \\
\left(1 + \frac{(m - m_{-})^2}{L_-}\right)^{-p_-} & \text{if } m < 0 
\end{cases}
\label{eq:Dm_param}
\end{equation}
where $m_{\pm}$ corresponds to the signed distance from the original activation point $P$ to an on-manifold intersection point $P_{\pm}$ along the steering line (Fig.~\ref{fig:mechanism}); 
$L_{\pm}$ sets the characteristic scale of decay and reflects how fast the distance-to-manifold grows along the steering direction (larger when the direction is locally parallel to the manifold and smaller when it cuts across it); 
and $p_{\pm}$ controls the decay rate (tail heaviness) as the trajectory moves away from the manifold neighborhood.

\subsection{Preference Capability: Projection Gain With Decay}
\label{subsec:pref-capability-projection}

We study how additive steering changes a model's preference through intermediate activations.
An intervention at layer $l$ updates the hidden state as $\tilde{h}(m)=h+m\,\Delta h$.

Prior work under LRH-style assumptions often models preference probability with a logistic form,
$P(p_p\mid h)=\sigma\!\big(-(\boldsymbol{\omega}_p^{\mathsf T}h+b_p)\big)$, where $\boldsymbol{\omega}_p$ is the preference vector.
Separately, work on activation geometry suggests that after low-dimensional projection (e.g., PCA), opposite preference labels are often approximately linearly separable. Under the activation-manifold view, this motivates a two-dimensional \emph{preference plane} and a preference direction whose signed coordinate reflects \emph{preference intensity}.
Our contribution is to incorporate validity attenuation $D(\cdot)$ (Assumption~\ref{ass:decay}) to account for off-manifold steering.

To model this, we write the steered preference probability as
\begin{equation}
P(p_p\mid \tilde{h}(m))
=\sigma\!\Big(-\big(\boldsymbol{\omega}_p^{\mathsf T}h+\alpha_p m\big)D_p(m)-b_p\Big),
\label{eq:pref_prob_steered}
\end{equation}
where $\alpha_p\triangleq \boldsymbol{\omega}_p^{\mathsf T}\Delta h$ measures how much the steering direction aligns with the preference vector: $\alpha$ is large when $\Delta h$ is aligned with $\boldsymbol{\omega}_p$, and $\alpha_p=0$ when $\Delta h$ is orthogonal to $\boldsymbol{\omega}_p$.

This implies the preference log-odds
\begin{equation}
\log\frac{P(p_p\mid \tilde{h}(m))}{1-P(p_p\mid \tilde{h}(m))}
=
\big(\boldsymbol{\omega}_p^{\mathsf T}h+\alpha_p m\big)D_p(m)+b_p.
\label{eq:pref_logodds_steered}
\end{equation}


\begin{thmbox}
\textbf{Key implication (linear regime $\rightarrow$ nonlinear collapse).}
From Eq.~\eqref{eq:pref_logodds_steered}, the $m$-dependence enters as $\alpha_p m\,D_p(m)$.
When $|m - m_{\pm}| \ll L_{\pm}$, Eq.~\eqref{eq:Dm_param} gives $D_p(m)\approx 1$, hence preference log-odds is approximately linear in $m$ with slope $\alpha$ (matching the near-linear regime in \citet{belief-dynamics}).
As $|m - m_{\pm}|$ grows and becomes comparable to or larger than $L_{\pm}$, Eq.~\eqref{eq:Dm_param} implies substantial decay in $D_p(m)$, so attenuation dominates and the log-odds response becomes strongly nonlinear and can collapse off-manifold.
\end{thmbox}

\paragraph{Fitting Form.}
We fit the measured preference log-odds as a function of $m$ with
\begin{equation}
\log\frac{P(p_p\mid \tilde{h}(m))}{1-P(p_p\mid \tilde{h}(m))}
\;=\;(\alpha_p m+\beta_p)\,D_p(m)\;+\;b_p,
\label{eq:logodds_fit}
\end{equation}
where $\beta_p=\boldsymbol{\omega}_p^{\mathsf T}\mathbf{h}$ is a per-example constant (since $\mathbf{h}$ is fixed for a given input), and $b_p$ is an offset. 

\paragraph{Fit Results.}
Table~\ref{tab:r2_scores} reports the fit quality of Eq.~\eqref{eq:logodds_fit}, with $R^2$ values exceeding 0.95 across most settings. These results validate the model's ability to accurately characterize the dynamics of preference log-odds.
Details are in Appendix~\ref{app:fit_details}.

\subsection{Utility Capability: Only Validity Decay}

\paragraph{Utility Log-odds Under Manifold-Validity Decay.}
Let $\mathbf{h}\in\mathbb{R}^{d_l}$ denote the activation at layer $l$.
We quantify utility capability by the log-odds of positive vs.\ negative utility outcomes ($u_p/u_n$).
Similar to preference, we assume utility is also associated with a direction $\boldsymbol{\omega}_u$ in activation space.
Under steering $\tilde{h}(m)=h+m\,\Delta h$, we model
\begin{equation}
\log\frac{P(u\mid \tilde{h}(m))}{1-P(u\mid \tilde{h}(m))} =
\boldsymbol{\omega}_u^{\mathsf T}\mathbf{h}\,D_u(m)+b_u,
\label{eq:util_logodds_decay}
\end{equation}
where $D_u(m)$ follows the manifold-validity decay in Eq.~\eqref{eq:Dm_param} and decreases with $|m|$.
Crucially, for \emph{preference} steering directions we typically have $\boldsymbol{\omega}_u^{\mathsf T}\Delta\mathbf{h}\approx 0$, so utility is affected primarily through validity decay rather than a direct projection term.

\paragraph{Fitting form.}
Accordingly, we fit the measured utility log-odds with a pure decay curve:
\begin{equation}
\log\frac{P(u\mid \tilde{h}(m))}{1-P(u\mid \tilde{h}(m))}
\;=\; \beta_u\,D_u(m)\;+\;b_u,
\label{eq:util_logodds_fit}
\end{equation}
where $\beta_u$ is the  baseline log-odds  and $b_u$ is an offset capturing residual bias.

\paragraph{Fit Results.}
Table~\ref{tab:r2_scores} reports the fit quality of Eq.~\eqref{eq:util_logodds_fit}. Uniformly high $R^2$ values (typically $>0.97$) suggest utility variations under preference steering are well captured by the proposed formulation. Additional details are in Appendix~\ref{app:fit_details}.

\begin{table}[!t]
\centering
\footnotesize
\setlength{\tabcolsep}{4pt}
\renewcommand{\arraystretch}{1.15}

\begin{tabular}{ll cccc cccc  @{\hspace{10pt}}  cccc cccc}
\toprule
& & \multicolumn{8}{c}{\textbf{Gemma-2-9B-IT}} & \multicolumn{8}{c}{\textbf{Qwen-2.5-7B-IT}} \\
\cmidrule(lr){3-10}\cmidrule(lr){11-18}
\multirow{2}{*}{\textbf{Type}} & \multirow{2}{*}{\textbf{Method}} &
\multicolumn{4}{c}{\textbf{Preference $R^2 \uparrow$}} & \multicolumn{4}{c}{\textbf{Utility $R^2 \uparrow$}} &
\multicolumn{4}{c}{\textbf{Preference $R^2 \uparrow$}} & \multicolumn{4}{c}{\textbf{Utility $R^2 \uparrow$}} \\
\cmidrule(lr){3-6}\cmidrule(lr){7-10}\cmidrule(lr){11-14}\cmidrule(lr){15-18}
& &
\textsc{Psy} & \textsc{Pwr} & \textsc{Axb} & \textbf{Avg} &
\textsc{Psy} & \textsc{Pwr} & \textsc{Axb} & \textbf{Avg} &
\textsc{Psy} & \textsc{Pwr} & \textsc{Axb} & \textbf{Avg} &
\textsc{Psy} & \textsc{Pwr} & \textsc{Axb} & \textbf{Avg} \\
\midrule

Weight & SFT  &
\cc 0.97 & \cb 0.98 & \ca 0.99 & \cb 0.98 &
\cb 0.98 & \ca 0.99 & \cb 0.98 & \cb 0.98 &
\ca 0.99 & \ca 0.99 & \ca 0.99 & \ca 0.99 &
\ca 0.99 & \ca 0.99 & \cb 0.98 & \ca 0.99 \\
       & RePS &
\ca 0.99 & \ca 0.99 & \ca 0.99 & \ca 0.99 &
\cc 0.96 & \ca 0.99 & \ca 0.99 & \cb 0.98 &
\ca 0.99 & \ca 0.99 & \ca 0.99 & \ca 0.99 &
\ca 0.99 & \ca 0.99 & \cc 0.97 & \cb 0.98 \\
\addlinespace

LoRA   & SFT  &
\cd 0.92 & \ca 0.99 & \cb 0.98 & \cc 0.96 &
\cb 0.98 & \ca 0.99 & \ca 0.99 & \ca 0.99 &
\cc 0.97 & \ca 0.99 & \ca 0.99 & \cb 0.98 &
\ca 0.99 & \ca 0.99 & \ca 0.99 & \ca 0.99 \\
       & RePS &
\cd 0.83 & \ca 0.99 & \ca 0.99 & \cd 0.94 &
\ca 0.99 & \ca 0.99 & \ca 0.99 & \ca 0.99 &
\cd 0.94 & \ca 0.99 & \ca 0.99 & \cc 0.97 &
\cb 0.98 & \cc 0.96 & \ca 0.99 & \cb 0.98 \\
\addlinespace

Vector & DiffMean &
\cc 0.97 & \ca 0.99 & \ca 0.99 & \cb 0.98 &
\cc 0.97 & \ca 0.99 & \cb 0.98 & \cb 0.98 &
\cb 0.98 & \cc 0.95 & \cb 0.98 & \cc 0.97 &
\ca 0.99 & \ca 0.99 & \cc 0.97 & \cb 0.98 \\
       & SFT &
\cd 0.93 & \cc 0.97 & \cb 0.98 & \cc 0.96 &
\ca 0.99 & \ca 0.99 & \ca 0.99 & \ca 0.99 &
\cd 0.93 & \cc 0.97 & \cb 0.98 & \cc 0.96 &
\cb 0.98 & \ca 0.99 & \ca 0.99 & \ca 0.99 \\
       & RePS &
\ca 0.99 & \cb 0.98 & \cc 0.95 & \cc 0.97 &
\ca 0.99 & \ca 0.99 & \ca 0.99 & \ca 0.99 &
\cc 0.97 & \cb 0.98 & \cd 0.93 & \cc 0.96 &
\ca 0.99 & \cb 0.98 & \ca 0.99 & \ca 0.99 \\

\bottomrule
\end{tabular}

\caption{\textbf{Curve fitting performance.} Results on Psychopathy (\textsc{Psy}), PowerSeeking (\textsc{Pwr}), and AXBench (\textsc{Axb}). We report $R^2$ (higher is better), measuring alignment between theoretical curves and empirical data. Color intensity indicates $R^2$ values. Consistently dark shading shows high fidelity across settings ($R^2 > 0.95$).}

\label{tab:r2_scores}
\end{table}
\section{Method}

\begin{table*}[t]
\centering
\small
\setlength{\tabcolsep}{6pt}
\resizebox{\textwidth}{!}{
\begin{tabular}{l l l c c cc}
\toprule
\multirow{3}{*}{Model} & \multirow{3}{*}{Form} & \multirow{3}{*}{Method} 
& Psychopathy 
& PowerSeeking 
& \multicolumn{2}{c}{AxBench} \\
\cmidrule(lr){4-4} \cmidrule(lr){5-5} \cmidrule(lr){6-7}
 &  & 
& Acc(\%, 0--100) $\uparrow$ 
& Concept(0--4) $\uparrow$ 
& Concept(0--2) $\uparrow$ 
& Harmonic(0--2) $\uparrow$ \\
\midrule

\multirow{13}{*}{Gemma-2-9B-IT}
& \multirow{1}{*}{Vanilla}
& Vanilla   & 50.00 & 1.87 & 0.4750 & 0.4950  \\
\cmidrule(lr){2-7}

& \multirow{3}{*}{Local Weight}
& SFT   & \bestcell{100.00} & \secondcell{3.50} & 1.6625 & 1.4538 \\
& 
& REPS  & \bestcell{100.00} & 3.39 & \secondcell{1.7750} & \bestcell{1.6362} \\
& 
& SPLIT (Ours)  & \bestcell{100.00} & \bestcell{3.59} & \bestcell{1.8500} & \secondcell{1.6225} \\
\cmidrule(lr){2-7}

& \multirow{3}{*}{LoRA}
& SFT   & \bestcell{100.00} & 3.41 & \secondcell{1.7625} & 1.5188 \\
& 
& REPS  & \secondcell{99.00} & \secondcell{3.44} & 1.7375 & \bestcell{1.6525} \\
& 
& SPLIT (Ours)  & \bestcell{100.00} & \bestcell{3.56} & \bestcell{1.7750} & \secondcell{1.6412} \\
\cmidrule(lr){2-7}

& \multirow{4}{*}{Vector}
& DiffMean & 53.00 & 2.95 & 1.1625 & 1.0550 \\
& 
& SFT        & 97.00 & 3.30 & \secondcell{1.7000} & 1.4487 \\
& 
& REPS       & \secondcell{98.00} & \secondcell{3.61} & \secondcell{1.7000} & \secondcell{1.5550} \\
& 
& SPLIT (Ours)       & \bestcell{99.00} & \bestcell{3.62} & \bestcell{1.8500} & \bestcell{1.6475} \\

\midrule

\multirow{13}{*}{Qwen-2.5-7B-IT}
& \multirow{1}{*}{Vanilla}
& Vanilla   & 50.00 & 2.24 & 0.4500 & 0.4713 \\
\cmidrule(lr){2-7}

& \multirow{3}{*}{Local Weight}
& SFT   & \secondcell{97.00} & \secondcell{3.53} & 1.5375 & 1.1287 \\
& 
& REPS  & 96.00 & 3.24 & \secondcell{1.6875} & \secondcell{1.4163} \\
& 
& SPLIT (Ours)  & \bestcell{98.00} & \bestcell{3.66} & \bestcell{1.7000} & \bestcell{1.4325} \\
\cmidrule(lr){2-7}

& \multirow{3}{*}{LoRA}
& SFT   & \secondcell{99.00}  & 3.05 & \secondcell{1.4875} & 1.3175 \\
& 
& REPS  & \bestcell{100.00}  & \secondcell{3.34} & \secondcell{1.4875} & \secondcell{1.4013} \\
& 
& SPLIT (Ours)  & \bestcell{100.00} & \bestcell{3.59} & \bestcell{1.7375} & \bestcell{1.6362} \\
\cmidrule(lr){2-7}

& \multirow{4}{*}{Vector}
& DiffMean & 55.00 & 3.17 & 0.9500 & 0.9950 \\
& 
& SFT        & \secondcell{97.00} & 3.58 & 1.5750 & 1.5800 \\
& 
& REPS       & 88.00  & \secondcell{3.63} & \secondcell{1.7375} & \secondcell{1.6412} \\
& 
& SPLIT (Ours)       & \bestcell{98.00} & \bestcell{3.65} & \bestcell{1.8125} & \bestcell{1.6500} \\

\bottomrule
\end{tabular}
}

\caption{
Main task performance of steering methods evaluated on three datasets.
\emph{Psychopathy} is reported with classification accuracy (Acc, \%), 
\emph{PowerSeeking} is evaluated using LLM-judge preference scores on a 0--4 scale, 
and \emph{AxBench} reports the concept score and the harmonic mean (HM) over concept, instruction, and fluency scores, each on a 0--2 scale, as evaluated by an LLM judge.
All methods perform inference-time interventions on hidden representations.
\bestcell{Best} and \secondcell{second-best} results are highlighted within each model and intervention form.
}
\vspace{1em}
\label{tab:control_form_comparison}
\end{table*}

\subsection{Preference--Utility Joint Optimization}

Building on the preceding mechanistic analysis, we propose \textbf{S}teering with \textbf{P}reference–Uti\textbf{L}ity \textbf{I}nterven\textbf{T}ion (SPLIT), a training objective improving preference while delaying utility degradation.

\paragraph{Utility Loss.}
To preserve utility, we train on both the positive and negative samples for the same input using the language-modeling cross-entropy:
\begin{equation}
\mathcal{L}_{\mathrm{util}}
= \lambda_p\,\mathcal{L}_{\mathrm{p}} + \lambda_n\,\mathcal{L}_{\mathrm{n}},
\end{equation}
where $\mathcal{L}_{\mathrm{p}}$ and $\mathcal{L}_{\mathrm{n}}$ are the token-level cross-entropy losses on positive and negative samples, respectively, and $\lambda_p,\lambda_n$ control their relative weight.

\paragraph{Preference Loss.}
By Eq.~\eqref{eq:pref-odds_main}, the loss gap $\mathcal{L}_{\mathrm{n}}-\mathcal{L}_{\mathrm{p}}$ is exactly the preference log-odds. We therefore maximize this gap via a hinge-style margin loss:
\begin{equation}
\mathcal{L}_{\mathrm{pref}}
= \gamma \cdot \sigma\!\left(\theta - (\mathcal{L}_{\mathrm{n}} - \mathcal{L}_{\mathrm{p}})\right),
\end{equation}
where $\sigma(\cdot)$ is ReLU, $\theta$ is a margin, and $\gamma$ balances preference improvement and utility preservation.

\paragraph{Final Objective.}
We combine the two components as
\begin{equation}
\mathcal{L}
= \mathcal{L}_{\mathrm{util}} + \mathcal{L}_{\mathrm{pref}}.
\end{equation}

\subsection{Experiment Results.}
\label{sec:our_experiment}
We evaluate the proposed preference-utility joint optimization method under three intervention forms: local weight update, low-rank adaptation (LoRA), and activation vector steering. 
As shown in Table~\ref{tab:control_form_comparison}, our approach consistently achieves higher scores  compared with baseline methods across all three intervention types.
These results demonstrate the robustness and generality of the proposed optimization strategy.
See Appendix~\ref{experiment} and~\ref{sec:dpo_results} for more experimental details and results.

\section{Related Work}
\label{app:related}
\paragraph{Mechanism.}
Most activation steering methods assume linear structure in activation space, controlling concepts by adding scaled direction vectors to hidden states~\citep{DBLP:conf/naacl/MikolovYZ13,DBLP:conf/emnlp/PenningtonSM14,DBLP:conf/blackboxnlp/NandaLW23,Tigges2023,DBLP:conf/icml/ParkCV24,wang2024knowledge,yao2025rethinking,zhang2026locate,hu2026towards}.
Building on this view, \citet{belief-dynamics} show that steering yields an approximately linear trend in posterior odds, but mainly in the small-scale regime.
Recent studies further report non-monotonic or adverse effects under stronger steering, challenging a naive global linearity assumption~\citep{bricken2023monosemanticity,DBLP:conf/icml/WollschlagerEGC25}. 
Meanwhile, representation-manifold work provides a complementary geometric lens for understanding steering and its limitations~\citep{origins-manifolds,li2025back,mhc}.

\paragraph{Activation Steering.}
Activation steering controls the behavior of LLMs by intervening in hidden states during forward propagation, using steering vectors to control single attributes as well as more complex behavioral targets~\citep{turner2023activation,CAA,DBLP:journals/corr/abs-2403-05767,DBLP:journals/corr/abs-2406-00244,DBLP:journals/corr/abs-2406-17563,DBLP:journals/corr/abs-2407-12404,DBLP:journals/corr/abs-2410-01174,DBLP:journals/corr/abs-2410-16314,DBLP:conf/eacl/KonenJDSBBOH24,DBLP:conf/emnlp/HazraL0P24,han2025internal,jiang2025anyedit}. 
However, recent studies have shown that the coarse-grained nature of activation steering can lead to degradation in model utility \citep{DBLP:conf/acl/WangXMDTC025,AxBench}.
\citet{bipo,DBLP:journals/corr/abs-2505-20809} introduce explicit preference learning objectives to optimize activation steering, enabling more precise control.

\paragraph{Parameter-Efficient Fine-Tuning.}
Parameter-Efficient Fine-Tuning (PEFT) methods, including adapters and LoRA, show that effective adaptation of LLMs does not require updating all parameters. 
LoRA achieves performance comparable to full fine-tuning, indicating that adaptation relies on structured low-rank weight updates rather than full parameter changes \citep{DBLP:conf/iclr/HuSWALWWC22,DBLP:journals/corr/abs-2309-02411,DBLP:conf/iclr/ZhangCBH0CZ23,DBLP:conf/icml/HayouG024,DBLP:conf/iclr/KopiczkoBA24,DBLP:conf/naacl/ZhangQSX24,WOS:001269604100009}. 
Local weight updates further reveal that LLM knowledge is highly localized, as modifying a small subset of parameters in specific layers suffices to change factual associations \citep{DBLP:conf/emnlp/GevaSBL21,DBLP:conf/acl/ZakenGR22,DBLP:journals/corr/abs-2203-06904,WOS:001356154700001,DBLP:journals/corr/abs-2509-22072}.


\section{Conclusion}
We propose a unified dynamic weight update framework that incorporates parameter updates, LoRA, and activation interventions, revealing a consistent preference–utility decay pattern in the log‑odds space. 
Building on this mechanistic insight, we design a joint optimization method that consistently improves preference while mitigating utility degradation across diverse intervention forms, demonstrating versatility and robustness.

\section*{Limitations}
While our unified dynamic weight update framework provides a coherent perspective on LLM control and enables predictable preference–utility trade-offs, several limitations remain. 
First, our analysis assumes that model representations lie near a well-structured activation manifold, which may not hold for extremely large or highly diverse models, potentially reducing the accuracy of our quantitative predictions. 
Second, our experiments focus primarily on attribute-level control (e.g., sentiment, style), leaving the applicability to complex multi-turn reasoning or safety-critical content largely unexplored. 
Third, while our proposed training objective mitigates the utility–preference trade-off, it does not guarantee complete avoidance of undesirable side effects such as subtle instruction violations or context drift under extreme control strengths.
Finally, our study evaluates control under pre-defined intervention multipliers, and generalization to adaptive or dynamically varying control signals requires further investigation.

\section*{Ethics Statement}
Controlled LLM generation carries inherent ethical considerations. 
While our framework aims to improve controllability and preserve task validity, it could potentially be misused to manipulate user perception, amplify biased viewpoints, or generate persuasive yet misleading content.
Our experiments are conducted on standard benchmark datasets and do not involve sensitive personal information. 
We emphasize that the proposed methods should be deployed with human oversight, adherence to fairness guidelines, and robust monitoring to prevent harm.
By explicitly modeling preference–utility trade-offs, we aim to make LLM interventions more interpretable and safer, but responsible usage depends on context-aware implementation and alignment with societal norms.

\section*{Acknowledgements}
We would like to express our sincere gratitude to the anonymous reviewers for their thoughtful and constructive feedback. This work was supported by the National Natural Science Foundation of China (No. 62576307, No. NSFCU23B2055, No. NSFCU19B2027), the Fundamental Research Funds for the Central Universities (226-2023-00138), the Yongjiang Talent Introduction Programme (2021A-156-G), and the Information Technology Center and State Key Lab of CAD\&CG, Zhejiang University.  
This work was supported by Alibaba Group through Alibaba Innovative Research Program.

\bibliographystyle{plainnat}
\bibliography{custom}
\clearpage
\appendix
\section*{Supplementary Material}
\section{Experiment Details}
\label{experiment}
\paragraph{Datasets.}
We evaluate our dynamic intervention methods on three datasets: (i) \emph{Psychopathy} (personality tendency classification), (ii) \emph{PowerSeeking} (open-ended generation), and (iii) the top-10 concept subsets from \emph{AxBench} (open-ended generation).
To support training and evaluation under our paired-setting and data availability constraints, we construct task-specific train/test splits as follows.
For \emph{Psychopathy}, we sample 500 instances for training and 100 for testing.
For \emph{PowerSeeking}, we sample 500 instances for training and 200 for testing.
For \emph{AxBench}, since its original test set is randomly sampled from an instruction-following corpus and does not provide matched positive/negative answer pairs, we re-split the original 72 instances per concept for each of the top-10 concept subsets into 64 training instances and 8 test instances.

\paragraph{Evaluation and Metrics.}
For the experiments in \S\ref{sec:unified_observation}, following~\citet{belief-dynamics}, we compute \textbf{preference} and \textbf{utility} log-odds (Eqs.~\eqref{eq:pref-odds_main} and~\eqref{eq:util-odds_main}) for each query $q$ with matched answers $(A_p, A_n)$ on both training and test sets, and vary the intervention scale $m$ to track their changes.
For the final performance evaluation in \S\ref{sec:our_experiment}, we adopt dataset-specific metrics.
For \emph{Psychopathy}, following~\citet{belief-dynamics}, we report classification accuracy (\textbf{Acc}).
For \emph{PowerSeeking}, following~\citet{bipo}, we use \texttt{gpt-4.1-mini} to score generations on the test set on a 0--4 scale.
For \emph{AxBench}, following~\citet{AxBench}, we use \texttt{gpt-4.1-mini} to evaluate \textbf{concept score}, \textbf{instruction score}, and \textbf{fluency score} on the test set, each on a 0--2 scale; we report the concept score and the \textbf{harmonic mean} over the three scores.

\paragraph{Baselines.}
We evaluate multiple methods under three intervention forms: \emph{local weight updates}, \emph{LoRA}, and \emph{vector} interventions. 
For each form, we train interventions with either the \textbf{SFT} objective or the \textbf{RePS} objective~\cite{AxBench}. 
For vector interventions, we additionally include a train-free baseline \textbf{DiffMean}~\cite{diffmean}. 
We also report \textbf{Vanilla} results without any steering.

\paragraph{Intervention Setup.}
We run experiments on \texttt{Gemma-2-9B-IT} at layer 20 and on \texttt{Qwen-2.5-7B-Instruct} at layer 14, following~\citet{belief-dynamics}. 
We consider three intervention forms: \emph{local weight updates}, \emph{LoRA}, and \emph{vector} interventions. 
For local weight and LoRA, we train intervention parameters on the MLP down-projection matrix; for vector interventions, we apply the intervention directly to the residual stream.
For hyperparameters, we largely follow the default settings in~\citet{AxBench, EasyEdit2}. 
We optimize with AdamW and a linear learning-rate scheduler. 
We also perform reasonable hyperparameter tuning to ensure stable and competitive performance.



\begin{figure*}[t]
    \centering
    \includegraphics[width=\linewidth]{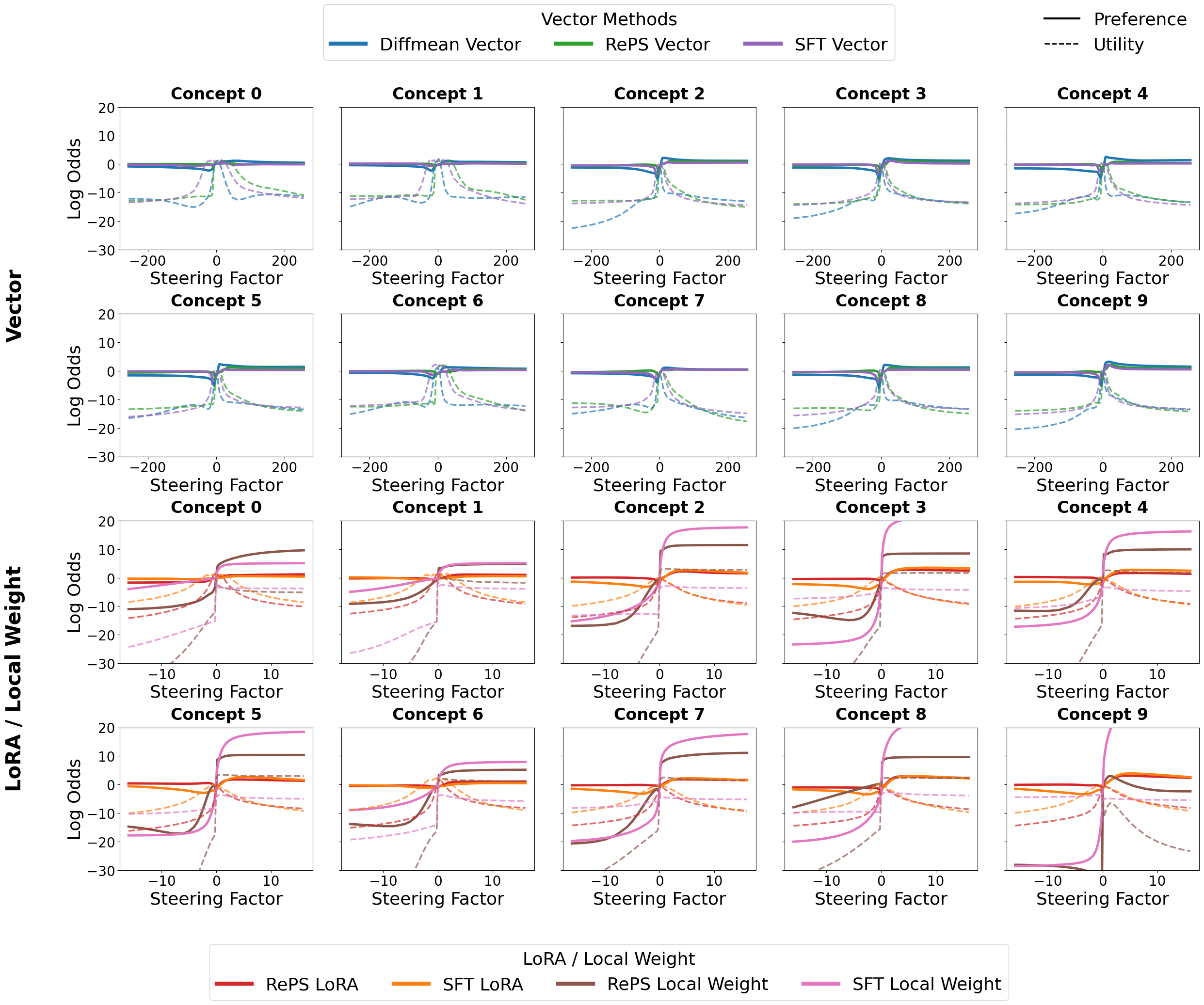}
    \caption{
    \textbf{Unified preference and utility dynamics under steering.}
        Solid lines represent preference log-odds, and dashed lines represent utility log-odds. 
        The top panel shows steering with vector-form parameter modifications, and the bottom panel shows parametric interventions including LoRA and local weight updates. 
        Results are shown for the \texttt{Qwen-2.5-7B-IT} model on the \emph{AxBench} dataset, evaluated over its top 10 concept subsets. 
        The horizontal axis corresponds to the steering factor.
    }
    \label{fig:unified-steering-qwen-it}
\end{figure*}

\begin{figure*}[!ht]
    \centering
    \begin{subfigure}[b]{0.9\linewidth} 
        \centering
        \includegraphics[width=\textwidth]{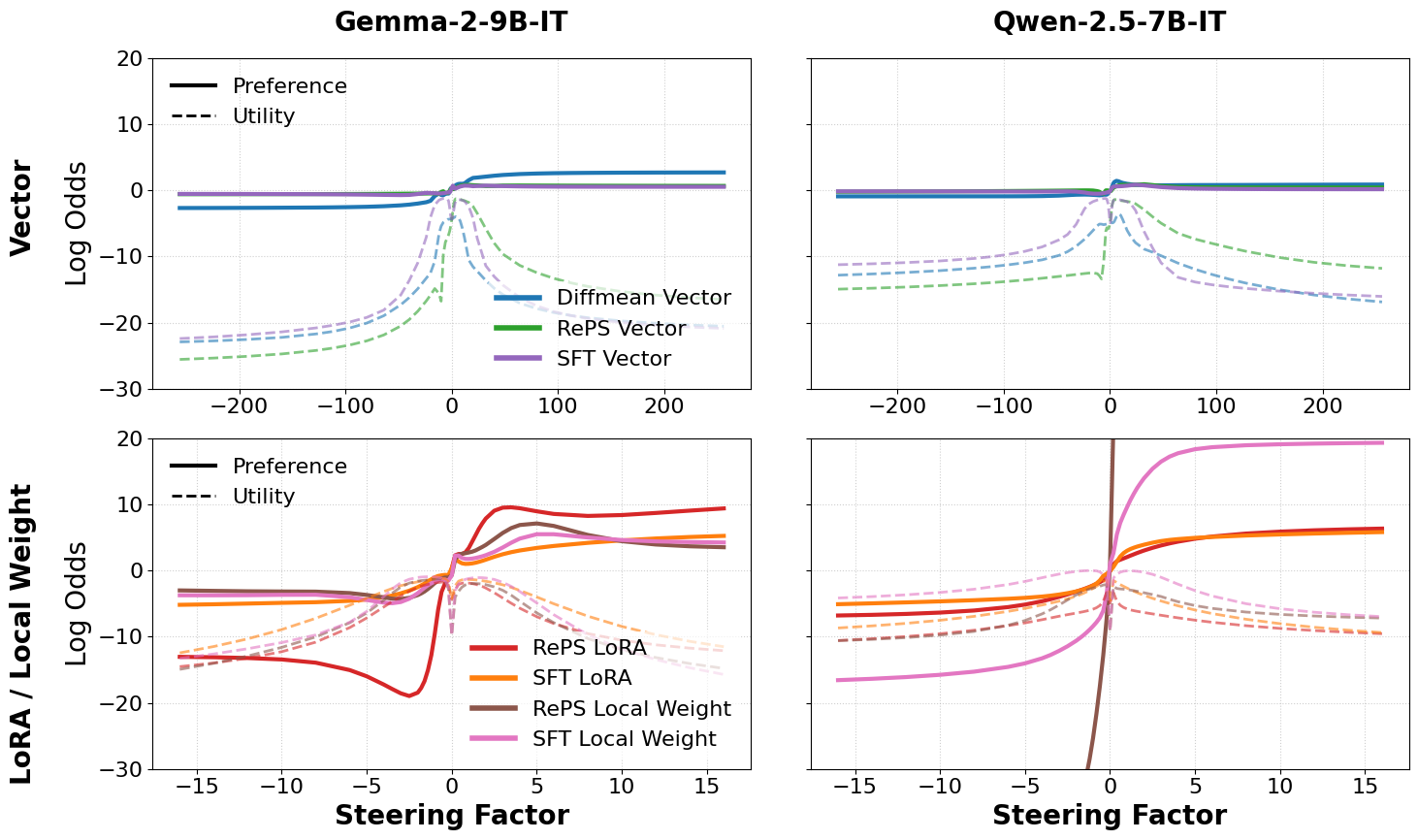}
        \caption{Powerseeking Results}
        \label{fig:sub1}
    \end{subfigure}
    
    \par\bigskip 
    
    \begin{subfigure}[b]{0.9\linewidth}
        \centering
        \includegraphics[width=\textwidth]{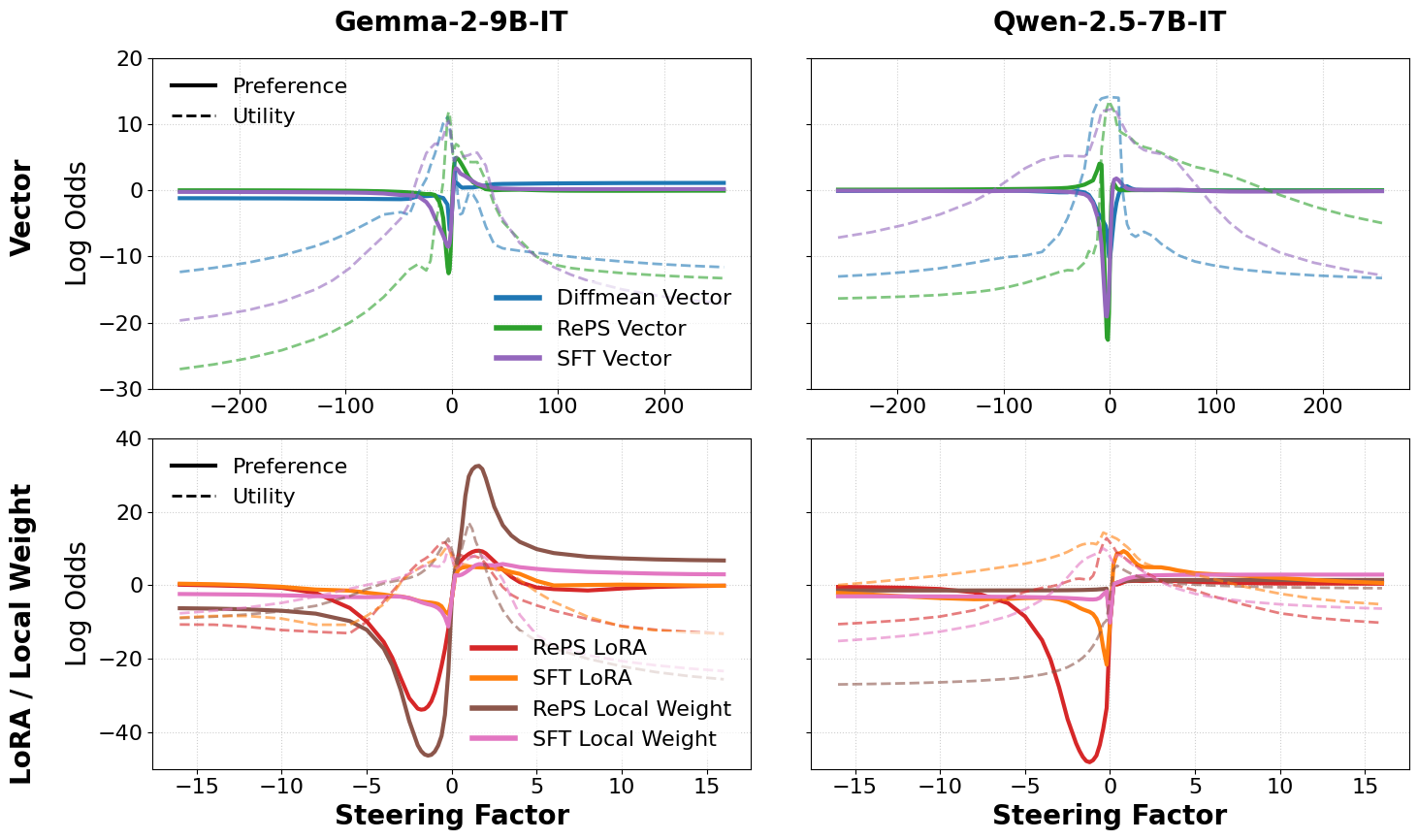}
        \caption{Psychopathy Results}
        \label{fig:sub2}
    \end{subfigure}
    
    \caption{
    \textbf{Unified preference and utility dynamics under steering.}
    Solid lines represent preference log-odds, and dashed lines represent utility log-odds. 
    Figure (a) shows the unified preference and utility dynamics of the power-seeking dataset under two different models, while Figure (b) shows the results for the psychopathy dataset.
    The horizontal axis corresponds to the steering factor.
    }
    \label{fig:main}
\end{figure*}

\paragraph{Results: Unified Dynamics Observation.}
Figures~\ref{fig:unified-steering} and~\ref{fig:unified-steering-qwen-it} show the unified preference–utility dynamics of the \texttt{Gemma-2-9B-IT} and \texttt{Qwen-2.5-7B-IT} models on the \emph{AxBench} dataset, evaluated over the top-10 concept subsets.
And figure \ref{fig:main} shows the unified preference-utility dynamics on \emph{Power-seeking} and \emph{Psychopathy} datasets under different models.
We observe that the utility can increase under slight perturbations of $m$ in either the positive or negative direction.
In some cases, this suggests that the origin may not lie exactly on the utility manifold, implying that the utility is not always strictly optimal at $m = 0$.

\paragraph{Results: Performance Comparison.}
Table~\ref{tab:control_form_comparison} compares our method with various baselines under different intervention forms (local weight, LoRA, and vector) on two base models.
Across intervention forms, our method remains competitive with strong baselines, and often improves concept metrics while maintaining comparable or higher harmonic scores.
The gains are most consistent under LoRA and vector, where our approach typically strengthens concept control relative to both SFT- and RePS-trained variants, and achieves the best or near-best harmonic mean on \textsc{AxBench} in multiple settings. 
Under full weight updates, we observe smaller but still stable differences, with our method remaining comparable and without an apparent drop in utility. 
Overall, the results indicate that the proposed optimization transfers across different steering forms and can provide reliable, albeit sometimes incremental, improvements.

\section{Results of DPO-based Method}
\label{sec:dpo_results}

\begin{table}[htbp]
\centering
\small
\resizebox{0.9\linewidth}{!}{
\begin{tabular}{lcccc}
\toprule
\textbf{Form} & \textbf{Psy.} & \textbf{Pow.} & \multicolumn{2}{c}{\textbf{AXB.}} \\
& \scriptsize{Acc} \scriptsize{(\%)}$\uparrow$ & \scriptsize{Concept} \scriptsize{(0--4)}$\uparrow$ & \scriptsize{Concept} \scriptsize{(0--2)}$\uparrow$ & \scriptsize{Harmonic} \scriptsize{(0--2)}$\uparrow$ \\
\midrule
Local Weight & 91.00 & 1.91 & 0.525 & 0.566 \\
LoRA         & 99.00 & 1.87 & 0.550 & 0.564 \\
Vector       & 99.00 & 1.93 & 0.575 & 0.631 \\
\bottomrule
\end{tabular}
}
\caption{Performance of the DPO-based method on Gemma-2-9B-IT across different adaptation forms. The evaluation encompasses Psychopathy (Psy.), PowerSeeking (Pow.), and AxBench (AXB.) metrics.}
\label{tab:dpo_comparison}
\end{table}

To provide a more exhaustive analysis of preference-based objectives in the context of behavior steering, we implemented a DPO-based method following BiPO~\citep{bipo}. 

As shown in Table~\ref{tab:dpo_comparison}, the DPO-based method performs consistently with prior observations reported in RePS~\citep{DBLP:journals/corr/abs-2505-20809}. 
In contrast, our method explicitly models the preference–utility trade-off, achieving a more balanced improvement across dimensions, indicating performance beyond a standard DPO-style approach.

\section{List of Mathematical Symbols}
The Table~\ref{tab:notations} below lists the important symbols used in this paper.

\begin{table}[!t]
    \centering
    \begin{tabular}{l p{11cm}} 
        \toprule
        \textbf{Symbol} & \textbf{Description} \\
        \midrule
        \multicolumn{2}{l}{\textit{\textbf{Unified Analysis Framework}}} \\
        $m$ & The steering scalar coefficient. \\
        $\text{PrefOdds}(q)$ & Preference log-odds ($\mathcal{L}_n - \mathcal{L}_p$). ~\eqref{eq:pref-odds_main} \\
        $\text{UtilOdds}(q)$ & Utility log-odds. ~\eqref{eq:util-odds_main} \\
        $P(u|q)$ & Latent utility probability. \\
        $P(p_{\pm}|q)$ & Latent preference probability. \\
        $P(\bullet |h)$ & Equivalent to $P(\bullet | q)$ as the weights remain unchanged. \\
        $\mathcal{L}_p, \mathcal{L}_n$ & Cross-entropy losses corresponding to $A_p$ and $A_n$. \\
        
        \midrule
        \multicolumn{2}{l}{\textit{\textbf{Mechanistic Manifold Model}}} \\
        $\mathcal{M}_l$ & The activation manifold of stably handled inputs at layer $l$. \\
        $D(m)$ & Average validity decay function. \\
        $m_{\pm}$ & Distance from P to P± along the steering line in Fig~\ref{fig:mechanism}\\
        $L_{\pm}$ & Characteristic scale of decay. \\
        $p_{\pm}$ & Asymptotic decay rate. \\
        
        \midrule
        \multicolumn{2}{l}{\textit{\textbf{Joint Optimization}}} \\
        $\mathcal{L}_{util}$ & Utility loss component. \\
        $\mathcal{L}_{pref}$ & Preference loss component. \\
        \bottomrule
    \end{tabular}
    \caption{\textbf{Notations for Key Concepts.} A summary of the specialized symbols introduced for the unified analysis, mechanistic modeling, and optimization objective.}
    \label{tab:notations}
\end{table}

\section{Derivations and Implementation Details for Log-Odds}
\label{app:logodds}

This appendix derives the loss-based forms of
Eqs.~\eqref{eq:pref-odds_main}--\eqref{eq:util-odds_main} from the
preference--utility independence assumption, and states how we compute the required
sequence losses.

\subsection{From preference--utility independence to log-odds}

For a query $q$ and a polarity pair $(A_p,A_n)$, we assume
\begin{align}
P(A_p\mid q) &= P(u\mid q)\,P(p_p\mid q), \nonumber\\
P(A_n\mid q) &= P(u\mid q)\,P(p_n\mid q),
\label{eq:app_factorization}
\end{align}
with $P(p_p\mid q)+P(p_n\mid q)=1$.

\paragraph{Preference log-odds.}
Taking the ratio of \eqref{eq:app_factorization} cancels $P(u\mid q)$:
\begin{align}
\frac{P(A_p\mid q)}{P(A_n\mid q)}
&=
\frac{P(p_p\mid q)}{P(p_n\mid q)}.
\label{eq:app_pref_ratio}
\end{align}
Applying $\log(\cdot)$ gives
\begin{align}
\mathrm{PrefOdds}(q)
&\triangleq
\log\frac{P(p_p\mid q)}{P(p_n\mid q)}
=
\log\frac{P(A_p\mid q)}{P(A_n\mid q)}.
\label{eq:app_prefodds_prob}
\end{align}
Using the loss definition $\mathcal{L}\triangleq -\log P(A\mid q)$, we have
$P(A\mid q)=e^{-\mathcal{L}}$, and thus
\begin{align}
\mathrm{PrefOdds}(q)
&=
\log\frac{e^{-\mathcal{L}_p}}{e^{-\mathcal{L}_n}}
=
\mathcal{L}_n-\mathcal{L}_p,
\label{eq:app_prefodds_loss}
\end{align}
which matches Eq.~\eqref{eq:pref-odds_main}.

\paragraph{Utility probability and log-odds.}
Summing \eqref{eq:app_factorization} and using $P(p_p\mid q)+P(p_n\mid q)=1$ yields
\begin{align}
P(A_p\mid q)+P(A_n\mid q)
&=
P(u\mid q) \nonumber\\
&\Big(P(p_p\mid q)+P(p_n\mid q)\Big)
\nonumber\\
&=
P(u\mid q).
\label{eq:app_u_prob}
\end{align}
Therefore,
\begin{align}
\mathrm{UtilOdds}(q)
&\triangleq
\log\frac{P(u\mid q)}{1-P(u\mid q)}
\nonumber\\
&=
\log\frac{P(A_p\mid q)+P(A_n\mid q)}
{1-P(A_p\mid q)-P(A_n\mid q)}.
\label{eq:app_utilodds_prob}
\end{align}
Substituting $P(A\mid q)=e^{-\mathcal{L}(A\mid q)}$ gives the loss form
\begin{align}
\mathrm{UtilOdds}(q)
&=
\log\frac{e^{-\mathcal{L}_p}+e^{-\mathcal{L}_n}}
{1-e^{-\mathcal{L}_p}-e^{-\mathcal{L}_n}},
\label{eq:app_utilodds_loss}
\end{align}
which matches Eq.~\eqref{eq:util-odds_main}.
Note that since $(A_p,A_n)$ are only two candidate continuations, we typically have
$P(A_p\mid q)+P(A_n\mid q)<1$.

\subsection{Computing sequence losses}

Let $A=(y_1,\dots,y_T)$ be a completion (excluding the query/prompt tokens).
We compute the sequence negative log-likelihood (cross-entropy loss) under teacher forcing:
\begin{align}
\mathcal{L}(A\mid q)
&\triangleq
-\log P(A\mid q)
\nonumber\\
&=
-\sum_{t=1}^{T}\log P(y_t \mid q, y_{<t}).
\end{align}
We then set $\mathcal{L}_p\triangleq \mathcal{L}(A_p\mid q)$ and
$\mathcal{L}_n\triangleq \mathcal{L}(A_n\mid q)$ and plug them into
Eqs.~\eqref{eq:app_prefodds_loss} and \eqref{eq:app_utilodds_loss}.

\paragraph{Length normalization (optional).}
When $A_p$ and $A_n$ have different lengths, we optionally use the mean loss
$\bar{\mathcal{L}}(A\mid q)\triangleq \mathcal{L}(A\mid q)/T$ in place of
$\mathcal{L}(A\mid q)$ to reduce length effects. In that case, the corresponding
quantities use $e^{-\bar{\mathcal{L}}}$ instead of $e^{-\mathcal{L}}$.

\subsection{Preference log-odds and Utility log-odds}

Here we show how the preference and utility capability can be represented as~\eqref{eq:pref_logodds_steered} and~\eqref{eq:util_logodds_decay}.
We take preference log-odds as example. First, the conditional probability before steering is given by:
\begin{align}
P(p_p\mid h)
&=\sigma\!\Big(-\boldsymbol{\omega}_p^{\mathsf T}hD_p(0)-b_p\Big)=
\sigma(\eta)
\end{align}
where $\eta \triangleq -\boldsymbol{\omega}_p^{\mathsf T}hD_p(0)-b_p$.

When an intervention at layer $l$ updates the hiden state as $\tilde{h}(m)=h+m\Delta h$, we can get the steered preference probablity as ~\eqref{eq:pref_prob_steered} :
\begin{equation}
\notag
P(p_p \mid \tilde{h}(m)) = \sigma\left( -(\boldsymbol{\omega}_p^\top h + \alpha_p m) D_p(m) - b_p \right)
\end{equation}

Next, we can represent the initial preference log-odds as $-\eta$:
\begin{align}
\log\frac{P(p_p\mid h)}{P(p_n\mid h)}
&=
\log\frac{P(p_p\mid h)}{1-P(p_n\mid h)} \notag\\
&=
\log\frac{\sigma(\eta)}{1-\sigma(\eta)} \notag\\
&=
\log\frac{1/(1+e^{\eta})}{1-1/(1+e^{\eta})} \notag\\
&=
\log\frac{1/(1+e^{\eta})}{e^{\eta}/(1+e^{\eta})} \notag\\
&=
\log e^{-\eta} \notag\\
&=
-\eta \notag\\
&= 
\boldsymbol{\omega}_p^{\mathsf T}hD_p(0)+b_p
\end{align}
Finally, when we steering $h$ by $\tilde{h}(m)$, we can get preference log-odds by:
\begin{align}
\log\frac{P(p_p\mid \tilde{h}(m))}{P(p_n\mid \tilde{h}(m))} 
&=
-\eta_{steered}=
(\boldsymbol{\omega}_p^\top h + \alpha_p m) D_p(m) + b_p
\end{align}

For utility capability, we have:
\begin{align}
P(u\mid \tilde{h}(m))
&=\sigma\!\Big(-\boldsymbol{\omega}_u^{\mathsf T} \tilde{h}(m)D_u(m)-b_u\Big).
\end{align}
For preference steering directions, we typically have $\boldsymbol{\omega}_u^{\mathsf T} \Delta h \approx 0$.
So we can quantify utility capability by:
\begin{equation*}
\log \frac{P(u \mid \tilde{h}(m))}{1 - P(u \mid \tilde{h}(m))} = \boldsymbol{\omega}_u^\top h D_u(m) + b_u
\end{equation*}

\section{Fitting Experiment Details}
\label{app:fit_details}

\subsection{Fitting Results on Test Set}
To further validate our theoretical model, we performed parameterized fitting on the test set using the SLSQP algorithm, strictly enforcing continuity between positive and negative segments at the origin. As shown in Table~\ref{tab:r2_scores_test}, the direct fitting yielded high goodness-of-fit values ($R^2 > 0.95$) for most methods. This confirms that the steering effect follows a deterministic trajectory predicted by our theory rather than random perturbations, thereby validating the proposed interaction mechanism.

\begin{table*}[t]
\centering
\setlength{\tabcolsep}{8pt} 
\renewcommand{\arraystretch}{1.1} 

\begin{tabular}{ll cccc cccc}
\toprule
\multirow{2}{*}{\textbf{Type}} & \multirow{2}{*}{\textbf{Method}} &
\multicolumn{4}{c}{\textbf{Preference $R^2 \uparrow$}} & \multicolumn{4}{c}{\textbf{Utility $R^2 \uparrow$}} \\
\cmidrule(lr){3-6}\cmidrule(lr){7-10}
& & \textsc{PSY} & \textsc{PWR} & \textsc{AXB} & \textbf{Avg} & \textsc{PSY} & \textsc{PWR} & \textsc{AXB} & \textbf{Avg} \\
\midrule

\multicolumn{10}{l}{\textit{\textbf{Gemma-2-9B-IT}}} \\
\midrule
Weight & SFT      & 0.96 & 0.96 & 0.99 & 0.97 & 0.98 & 0.93 & 0.99 & 0.97 \\
       & RePS     & 0.95 & 0.98 & 0.95 & 0.96 & 0.98 & 0.93 & 0.99 & 0.97 \\
\addlinespace 
LoRA   & SFT      & 0.99 & 0.99 & 0.98 & 0.99 & 0.98 & 0.98 & 0.99 & 0.98 \\
       & RePS     & 0.99 & 0.99 & 0.98 & 0.99 & 0.99 & 0.99 & 0.99 & 0.99 \\
\addlinespace
Vector & DiffMean & 0.89 & 0.99 & 0.99 & 0.96 & 0.94 & 0.99 & 0.98 & 0.97 \\
       & SFT      & 0.90 & 0.97 & 0.97 & 0.95 & 0.98 & 0.99 & 0.99 & 0.99 \\
       & RePS     & 0.96 & 0.98 & 0.96 & 0.97 & 0.96 & 0.99 & 0.99 & 0.98 \\

\midrule

\multicolumn{10}{l}{\textit{\textbf{Qwen-2.5-7B-IT}}} \\
\midrule
Weight & SFT      & 0.99 & 0.82 & 0.99 & 0.93 & 0.99 & 0.99 & 0.95 & 0.98 \\
       & RePS     & 0.99 & 0.89 & 0.97 & 0.95 & 0.98 & 0.99 & 0.90 & 0.96 \\
\addlinespace
LoRA   & SFT      & 0.70 & 0.95 & 0.98 & 0.88 & 0.99 & 0.99 & 0.99 & 0.99 \\
       & RePS     & 0.88 & 0.95 & 0.95 & 0.93 & 0.98 & 0.99 & 0.98 & 0.98 \\
\addlinespace
Vector & DiffMean & 0.99 & 0.99 & 0.98 & 0.99 & 0.97 & 0.94 & 0.98 & 0.96 \\
       & SFT      & 0.99 & 0.99 & 0.97 & 0.98 & 0.97 & 0.95 & 0.99 & 0.97 \\
       & RePS     & 0.99 & 0.98 & 0.93 & 0.97 & 0.96 & 0.96 & 0.98 & 0.97 \\
\bottomrule
\end{tabular}
\caption{\textbf{Performance comparison of curve fitting quality on test sets.} We evaluate the models on three datasets: Psychopathy (\textsc{PSY}), PowerSeeking (\textsc{PWR}), and AXBench (\textsc{AXB}). }
\label{tab:r2_scores_test}
\end{table*}

\subsection{Analysis of Generalization Ability}
Following the validation of our theoretical mechanism, we conducted train-to-test transfer experiments to evaluate the extent to which different methods decouple "concepts" from specific inputs. Theoretical curve parameters were derived solely from training data and applied directly to the test set for prediction (Table~\ref{tab:generalization_r2}).

\paragraph{Robust Generalization} 
Overall, the fitted curves generalize well to held-out data, with vector-based interventions achieving consistently strong $R^2$ across most settings. 
Input-dependent approaches such as LoRA- and local-weight-based methods also generalize well in many cases, but exhibit larger variance across datasets and occasional failures, suggesting that input-dependent updates can be more sensitive to the evaluation distribution.


\begin{table}[htbp]
\centering
\setlength{\tabcolsep}{8pt} 
\renewcommand{\arraystretch}{1.1} 

\begin{tabular}{ll cccc cccc}
\toprule
\multirow{2}{*}{\textbf{Type}} & \multirow{2}{*}{\textbf{Method}} &
\multicolumn{4}{c}{\textbf{Preference $R^2 \uparrow$}} & \multicolumn{4}{c}{\textbf{Utility $R^2 \uparrow$}} \\
\cmidrule(lr){3-6}\cmidrule(lr){7-10}
& & \textsc{PSY} & \textsc{PWR} & \textsc{AXB} & \textbf{Avg} & \textsc{PSY} & \textsc{PWR} & \textsc{AXB} & \textbf{Avg} \\
\midrule

\multicolumn{10}{l}{\textit{\textbf{Gemma-2-9B-IT}}} \\
\midrule
Weight & SFT      & 0.96 & 0.85 & -4.25 & -0.81 & 0.98 & 0.98 & 0.61 & 0.86 \\
       & RePS     & 0.99 & 0.98 & -1.16 & 0.27  & 0.96 & 0.98 & 0.73 & 0.89 \\
\addlinespace
LoRA   & SFT      & 0.92 & 0.99 & -0.56 & 0.45  & 0.98 & 0.99 & 0.96 & 0.98 \\
       & RePS     & 0.83 & 0.99 & 0.74  & 0.85  & 0.98 & 0.99 & 0.97 & 0.98 \\
\addlinespace
Vector & DiffMean & -0.14& 0.99 & 0.75  & 0.53  & 0.97 & 0.99 & 0.97 & 0.98 \\
       & SFT      & 0.90 & 0.91 & 0.74  & 0.85  & 0.98 & 0.99 & 0.99 & 0.99 \\
       & RePS     & 0.98 & 0.89 & 0.65  & 0.84  & 0.99 & 0.99 & 0.99 & 0.99 \\

\midrule

\multicolumn{10}{l}{\textit{\textbf{Qwen-2.5-7B-IT}}} \\
\midrule
Weight & SFT      & 0.99 & -0.32 & -12.03& -3.79 & 0.99 & -1.33& -3.07& -1.14\\
       & RePS     & 0.96 & 0.98  & -3.82 & -0.63 & 0.99 & 0.42 & -1.15& 0.09 \\
\addlinespace
LoRA   & SFT      & 0.97 & 0.98  & -0.40 & 0.52  & 0.99 & 0.99 & 0.95 & 0.98 \\
       & RePS     & 0.94 & 0.99  & -0.13 & 0.60  & 0.98 & 0.96 & 0.96 & 0.97 \\
\addlinespace
Vector & DiffMean & 0.86 & 0.94  & 0.80  & 0.87  & 0.37 & 0.99 & 0.97 & 0.78 \\
       & SFT      & 0.67 & 0.92  & 0.71  & 0.77  & 0.96 & 0.99 & 0.99 & 0.98 \\
       & RePS     & 0.97 & 0.93  & 0.74  & 0.88  & 0.99 & 0.98 & 0.98 & 0.98 \\
\bottomrule
\end{tabular}
\caption{\textbf{Generalization ability of curve fitting.} The table reports the $R^2$ scores where the curves are fitted on the training set and evaluated on the test set across three datasets: Psychopathy (\textsc{PSY}), PowerSeeking (\textsc{PWR}), and AXBench (\textsc{AXB}). Negative values imply that the fitted curves do not generalize well to unseen data.}
\label{tab:generalization_r2}
\end{table}

\end{document}